\documentclass[10pt,twocolumn,letterpaper]{article}

\usepackage{cvpr}
\usepackage{times}
\usepackage{epsfig}
\usepackage{graphicx}
\usepackage{amsmath}
\usepackage{stmaryrd}
\usepackage{amssymb}
\usepackage{multirow}
\usepackage{verbatim}
\usepackage{color}
\usepackage{float}
\usepackage{enumitem}
\usepackage{booktabs}
\usepackage{subcaption}
\usepackage{makecell}
\usepackage{soul}
\usepackage{mathtools}
\usepackage{epigraph}

\usepackage{amssymb}
\usepackage{times}
\usepackage{epsfig}
\usepackage{graphicx}
\usepackage{amsmath}
\usepackage{stmaryrd}
\usepackage{amssymb}
\usepackage{multirow}
\usepackage{verbatim}
\usepackage{color}
\usepackage{float}
\usepackage{enumitem}
\usepackage{booktabs}
\usepackage{subcaption}
\usepackage{makecell}
\usepackage{soul}
\usepackage{mathtools}
\usepackage{epigraph}
\usepackage{bbm}

\usepackage{mathbbol}

\usepackage[pagebackref=true,breaklinks=true,letterpaper=true,colorlinks,bookmarks=false]{hyperref}
\hypersetup{
	plainpages=false,
	colorlinks=true,              %
	linkcolor=green,               %
	anchorcolor=blue,             %
	citecolor=green,               %
	filecolor=blue,               %
	urlcolor=blue,                %
	pdfview=FitH,                 %
	pdfstartview=FitH,            %
	pdfpagelayout=SinglePage      %
}

\cvprfinalcopy 
\usepackage{pifont}
\def\cvprPaperID{2151} 

\ifcvprfinal\fi
\begin{document}


\title{Weak Supervision and Referring Attention for Temporal-Textual Association Learning}
\author{
	Zhiyuan Fang\textsuperscript{1}\footnotemark[1]
	\quad\quad\quad
	Shu Kong\textsuperscript{2}\footnotemark[1]
	\quad\quad\quad
	Zhe Wang\textsuperscript{3}\footnotemark[1]
	\\
		\quad\quad
	Charless Fowlkes\textsuperscript{4}
	\quad\quad\quad\quad\quad
	 Yezhou Yang\textsuperscript{1}
	\\
	\small{$^1$Arizona State University \quad $^2$\normalsize{Carnegie Mellon University} \small \quad $^3$Beihang University \quad $^4$University of California, Irvine}
		}

\pagestyle{plain}

\twocolumn[{%
\renewcommand\twocolumn[1][]{#1}%
\vspace{-0.5cm}
\maketitle

\vspace{-3em}
\begin{center}
\includegraphics[width=\linewidth]{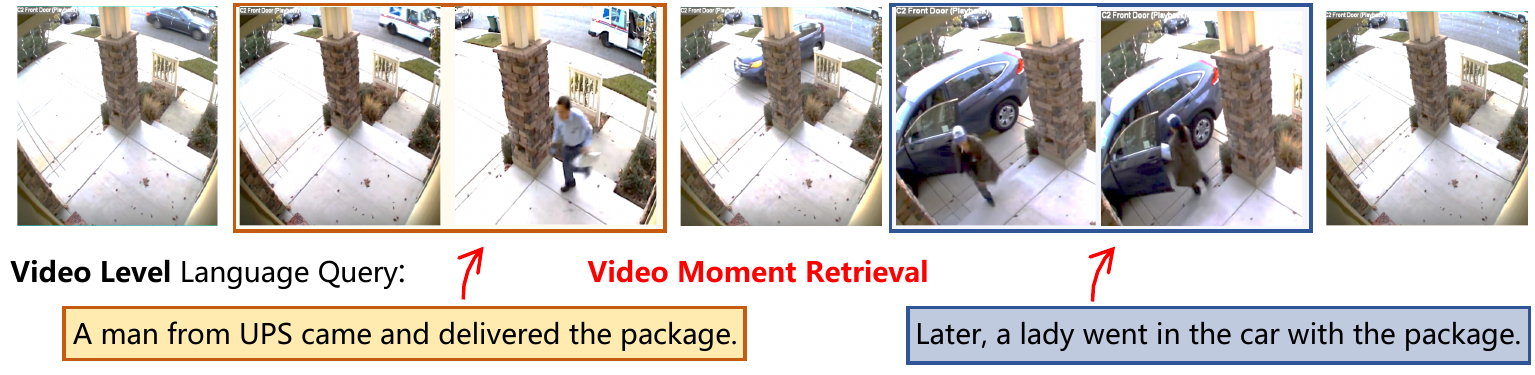} 
\captionof{figure}{	\small 
One practical application of the weakly supervised temporal textual association learning can be video moment retrieval (\textit{e.g.}, in surveillance video) using natural language query, which requires the video-level caption to align with the video segment temporally without any annotation.   
}
\vspace{-2mm}

\label{fig:abstract}
\end{center}%
}]

\begin{abstract}
\vspace{-2mm}
A system capturing the association between video frames
and textual queries offer great potential for better video analysis.
However, training such a system in a fully supervised way inevitably demands 
a  meticulously curated video dataset with temporal-textual annotations. 
Therefore we provide a Weak-Supervised alternative with
our proposed Referring Attention mechanism to learn temporal-textual association (dubbed \emph{WSRA}).
The weak supervision is simply a textual expression (\textit{e.g.}, short phrases or sentences) at video level,
indicating this video contains relevant frames.
The referring attention is our designed mechanism acting as a scoring
function for grounding the given queries over frames temporally.
It consists of multiple novel losses and sampling strategies for better training.
The principle in our designed mechanism is to fully exploit 1) the weak supervision by considering informative and discriminative cues from
intra-video segments anchored with the textual query, 2) multiple queries compared to the single video, and 3) cross-video visual similarities.
We validate our \textit{WSRA} through extensive experiments for temporally grounding by languages, demonstrating that it outperforms the state-of-the-art weakly-supervised methods notably.
\end{abstract}
\vspace{-4mm}

\section{Introduction}
Videos 
contain much richer information for humans to interpret the world.
Building an intelligent system to automate video analysis could yield a wide range of applications benefiting human society at large, from assistive robots for the elderly to video surveillance for security~\cite{kim2018textual,yang2015robot,oh2011large,fang2020video2commonsense}.
A significant component in such a system is to capture the 
association between video frames and textual reference/queries,
\emph{a.k.a} temporal-textual association~\cite{gao2017tall,anne2017localizing,liu2018cross,ge2019mac}.
Therefore, learning temporal-textual association over videos becomes a promising direction in the community~\cite{anne2017localizing,sun2019videobert,miech2019howto100m}.

One way to obtain such a model for temporal-textual association learning 
is to fully supervise the training over a video dataset which has 
meticulous annotations in term of the associated
frames with some textual queries~\cite{chen2018temporally,anne2017localizing,gao2017tall,zhang2019man,ge2019mac}.
However, we note that such a practice inevitably demands
a large-scale  dataset,
which apparently is not only prohibitively expensive to collect, 
but also largely limited in terms of diversity of both videos and  textual expressions. 
As an alternative,
a few recent methods propose to learn the temporal-textual association 
only with weak annotations, \emph{i.e.}, 
video-level expressions in the form of natural language description~\cite{Mithun_2019_CVPR,lin2019weakly}.

Though the weakly-supervised temporal-textual association learning
attracted increasing attention until just recently, there exists a great number of works on related topics, such as weakly supervised action localization in videos~\cite{singh2017hide,paul2018w,mishra2018generative,nguyen2018weakly}.
However, compared to action localization which only has a limited number of action categories,
grounding textual reference is more challenging since the textual expressions 
could be more free-form with multiple words for the same meaning and flexible sentence structures.
In another word,
using the natural-language descriptions greatly
enlarges the content and diversity of visual-expression searching;
the combinatorial nature of open-form languages also makes it infeasible 
to enumerate all possible expressions towards the same indication.
For instance, 
instead of just localizing the video frames with categorical labels like  ``\textit{kissing}'' or ``\textit{person}'', 
a more practical and user-friendly query might be 
``\textit{the moment when the new couple are kissing in the wedding}'' (for moment retrieval) 
or ``\textit{a man in yellow shirt appearing in the hall in last night}'' (for video surveillance).
This necessitates the study of temporal grounding using natural language descriptions.
To foster the study, 
Gao \emph{et al.} augment the Charades 
dataset~\cite{sigurdsson2016hollywood} by generating complex language queries with temporal boundary annotations~\cite{gao2017tall};
Anne Hendricks \emph{et al.} collect the DiDeMo dataset with manual annotations 
on the associated video frames and natural-language descriptions~\cite{anne2017localizing}.
Both datasets for the first place enable training for language moment retrieval or temporal grounding. 

In this paper, basing on the datasets available in the literature,
we study learning temporal-textual association with 
weak supervisions, \emph{e.g.}, video-level language expressions
as shown in Fig.~\ref{fig:abstract}.
Until very recently, few works propose to weakly supervised train for temporal-textual association with language expressions.
In particular,
Mithun \emph{et al.} present the very first attempt for weakly supervised training
to localize video segment over the given textual queries~\cite{Mithun_2019_CVPR}. 
In general, weakly supervised methods of temporal-textual association learning
are facing two major challenges: 
1) the lack of precise supervision aligning video segments and textual queries,
2) highly undecidable features for the complex and open-form languages.



To overcome these challenges, 
we propose a weakly-supervised framework with a referring attention mechanism (\textit{WSRA})
for learning temporal-textual associations on videos.
The proposed referring attention mechanism summarizes a series of our novel components.
The first one learns through a background modelling method to pool out 
irrelevant frames specific to the given language query. 
Building upon it, 
the second component encourages foreground features to align with the query
by discriminating itself from the background features from the first component. 
Within this component, 
we present a hard negative mining method integrated to sample irrelevant textual descriptions during learning. 
The third component exploits inter-video (dis-)similarity based on 
the multiple textual queries.
Specifically,
this forces the visual features to be close to each other from different videos,
as long as the queries convey similar meanings measured by the similarity of textual features.

To summarize our  contributions:
1) We propose a unified framework (\textit{WSRA}) for weakly supervised learning temporal-textual associations with the referring attention 
mechanism,
directly applicable to moment retrieval and language grounding in videos.
2) We show with rigorous ablation study that the proposed components 
in the referring attention leverages better informative cues from  the limited weak supervision.
3) We justify the \textit{WSRA} through extensive experiments,
notably outperforming other state-of-the-art weakly-supervised methods on these
tasks on two public benchmarks, 
DiDeMo \cite{anne2017localizing} and 
Charades-STA \cite{gao2017tall}.

\section{Related Work}

\noindent{\bf Association Learning across Vision and Language} is core and tie of a wide range of tasks across vision and language domains, \emph{e.g.}, textual grounding~\cite{plummer2015flickr30k}, referring expression comprehension~\cite{nagaraja2016modeling} or object retrieval using language~\cite{hu2016natural}. Recent works focus on leveraging the image-level annotations (as weak supervision)~\cite{fang2018modularizedtextual,fang2018weakly} or unsupervised method~\cite{yeh2018unsupervised} to learn the association across language descriptions and objects. Proceeding from this, there arise works on using uncurated captions to learn temporal associations across video segments and texts~\cite{miech2019end,sun2019videobert}. Notably, these works all highlight on the importance of constructing contrastive pairs in exploiting the weak annotations and inspire our work.

\noindent{\bf Weakly-Supervised Action Localization} 
can be thought of a specific example of weakly supervised video learning
with the video-level labels~\cite{sun2015temporal,shou2018autoloc,nguyen2018weakly,paul2018w}.
This problem derives from  
the fully-supervised counterpart methods which exploits fine annotations at frame level
for localizing the actions~\cite{buch2017sst,kalogeiton2017action,weinzaepfel2015learning,shou2017cdc,tran2012max,shou2016temporal}.
Recent weakly-supervised methods extensively adopt either the video-level classification framework~\cite{singh2017hide,shou2018autoloc,sikka2014classification,dwibedi2019temporal} or with the attentional mechanism 
that generates bottom-up sparse weights used for localizing action categories temporally~\cite{nguyen2018weakly}. Beyond that, few works~\cite{wang2017untrimmednets,paul2018w} propose to utilize the multi-instance learning loss to address this challenge,~\cite{paul2018w} also suggests  exploiting the co-activity across videos in the metric learning, which largely improves the action localization task even when temporal annotations are not available. Most recent work~\cite{nguyen2019weakly} improves over these methods with background modelling module that explicitly extract the foreground and background appearances. 

\noindent{\bf Temporal Grounding and Moments Retrieval}
are two instantiations of learning  
temporal-textual association.
In these tasks,
most recent methods adopt fully-supervised training over 
fine annotations on the frame-textual associations.
For example,
Gao \emph{et al.} augment the Charades 
dataset~\cite{sigurdsson2016hollywood} by generating complex language queries with temporal boundary annotations for language moment retrieval~\cite{gao2017tall}; 
Anne Hendricks \emph{et al.} also collect a new dataset for training to localize video moments over a given descriptive sentence~\cite{anne2017localizing}. 
Other follow-up methods~\cite{hendricks2018localizing,liu2018attentive,chen2018temporally,wang2018bidirectional,xu2019joint,gavrilyuk2018actor} also
fully-supervised train for temporal grounding using these datasets,
suffering from the limitation on their generalizability due to  
the combinatorial nature of complex natural language sentences (\textit{e.g.}, 
synonymous words, grammatical tense 
and sentence structures)~\cite{hendricks2018localizing,gao2017tall}. 

Weakly-Supervised Video Grounding by Language further 
advances the aforementioned to an even challenging task, as now the only available supervision is video-level natural language descriptions in open format, which come with huge amount of unnecessary noises.
In particular, Mithun \emph{et al.}~\cite{Mithun_2019_CVPR} for the first time attempt to 
solve the temporal grounding problem by weakly supervised learning 
with only the video-level textual queries~\cite{Mithun_2019_CVPR}. 
In~\cite{chen2020look}, the author proposed to utilize the temporal proposal and textual description alignment learning using the sliding window fashion, and tackle the grounding problem in a coarse-to-fine manner. Similarly, a very recent work by~\cite{lin2019weakly} also focuses on the design of a better proposal generation module, that aggregates the contextual visual cues to generate and score the proposal candidates for grounding. We summarize that current efforts in weakly supervised language grounding are either centered on better proposal generation~\cite{chen2020look,lin2019weakly}, or a better cross-modal association model by constructing contrastive samples across video and languages~\cite{Mithun_2019_CVPR,gao2019wslln}. Our WSRA places emphasis on the latter, but nevertheless further distinguishes the above with a more comprehensive cross-modal association learning objectiveness, and a novel sampling and weighting strategy in our metric learning step.


\section{Temporal-Textual Association Learning}
\begin{figure}[t]
\centering
\includegraphics[width=.49\textwidth, height=.13\textwidth]{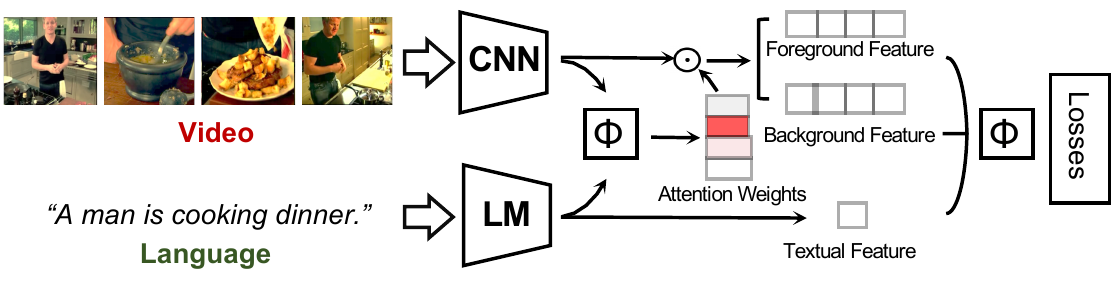}
\caption{\small Flowchart of the proposed \textit{WSRA}.
The losses are only used during training for temporal-textual association learning,
while inference shares the same computation flow 
for grounding textual queries over the given video. ``\textit{CNN}'' and ``\textit{LM}'' denote
video model and language model to extract features for frames and the textual query.
respectively. $\Phi$ denotes the cross-modal scoring function, and the above shown two functions are with independent learning parameters.
}
\label{figure:over_arch}
\end{figure}

The well learned temporal-textual associations on videos can allow for practical tasks like moment retrieval and temporal grounding of natural-language descriptions, where the core to it is to learn a joint embedding space for both the frames and the textual queries.
In this embedding space, the associated frames and the textual queries should be close to each other represented by the visual feature and language feature, respectively.
Formally, given a long and untrimmed video ${\cal V}=[\boldsymbol{v}_1, ..., \boldsymbol{v}_T]$ consisting of $T$ snippets of frames, and an open-form textual sentence $\boldsymbol{t}$, we would like to train a model to localize a video segment $\boldsymbol{v}_t$ from ${\cal V}$ that best corresponds to the description, which is ideally the true (yet unknown) video segment for the textual query.
We denote by $\boldsymbol{v}_t \in\mathbb{R}^{d}$ the feature vector 
extracted from a video model (\emph{e.g.}, a pretrained classification network),
and by $\boldsymbol{t}\in\mathbb{R}^{d}$ the textual feature representation of the 
query (short phrase or a sentence) from a language model.
In practice, segment features can also be the averagely pooled proposal visual features, where each proposal contains several continuous segments with various lengths. 
As weak supervision causes ambiguities in predicting the association between frames and the textual query, we present our weakly-supervised approach with the proposed referring attention mechanism (\textit{WSRA}) in this section.
Fig.~\ref{figure:over_arch} shows the overall architecture of our model.

\begin{figure*}[t]
\centering
\includegraphics[width=.95\textwidth]{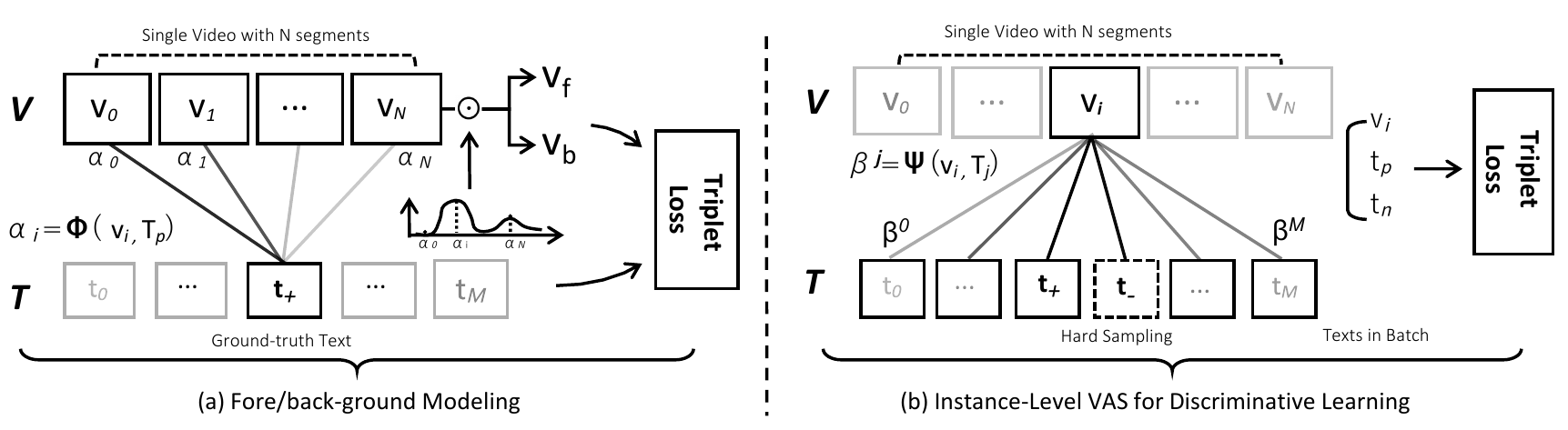}
\caption{\small The \textit{WSRA} framework contains three learning losses. (a) Fore/background modelling loss forces the fore/background visual features to be discriminative by training 
over the same video \textit{w.r.t.} the ground-truth referring text $\mathbf{t}_\textit{$+$}$. (b) Our snippet-level attention modeling samples negative textual queries $\mathbf{t}_\textit{$-$}$ for each video segment for discriminative learning. The third loss is shown separately in Fig.~\ref{figure:crossvideo}.
}
\label{figure:architecture}
\end{figure*}

\subsection{Video-level Attention Modeling}
Prior weakly supervised methods for action localization propose to generate a weight vector to localize action labels among the video snippets in a \textit{bottom-up} manner~\cite{paul2018w,nguyen2019weakly}. While being successful in action localization from predefined categorical labels, these bottom-up methods cannot be directly tailored to the more natural language localization problem in question, as a video can correspond to multiple textual queries which appear in a very open-form language structure. Based on the fact that, the description of textual query usually covers multiple snippets in a video, we propose to model the video-level back/fore-ground features that are (ir)related to the textual query with the designed referring weights:
\begin{equation}
\begin{split}
    \boldsymbol{v}_f = &\sum^{T}_{t=1}\alpha_t\boldsymbol{v}_t \\
    \boldsymbol{v}_b = \frac{1}{T-1} &\sum^{T}_{t=1}(1-\alpha_t)\boldsymbol{v}_t,
\end{split}
\label{eq:foreback}
\end{equation}
where $v_f$ and $v_b$ are the synthesized fore/back-ground features, and weight $\alpha_t$ is calculated as:
\begin{equation}
    \alpha_t = \frac{\text{exp}(\phi(\boldsymbol{v}_t, \boldsymbol{t}))}{{\sum_{i=1}^{T}\text{exp}(\phi(\boldsymbol{v}_i, \boldsymbol{t}))}}.
    \label{eq:videoweight}
\end{equation}
In Eq.~\ref{eq:videoweight}, $\phi(\cdot, \cdot)$ is the cross-modal scoring function that measuring the distance between a frame/snippet and the textual query in the embedding space.
In practice, we define $\phi(\boldsymbol{v}, \boldsymbol{t})=\text{Sigmoid}(\text{FC}(\text{Cat}(\boldsymbol{v}, \boldsymbol{t}))$ with learnable parameters, which has a stronger association ability than cosine similarity~\cite{yu2018mattnet}, bilinear pooling~\cite{gao2016compact,fang2018modularizedtextual,kong2017low}, and second-order polynomial feature fusion~\cite{scikit-learn,gao2017tall}.
It is worth noting that the scoring function not only conveys the idea of metric learning, 
but also lands for embedding learning along with further loss terms as presented in next subsections.

As we would like to learn an embedding space in which,
1) foreground visual features tightly correspond to the given textual query measured by
the scoring function performing in the same space;
and 2) background features are clearly far away from both the foreground and textual query, as illustrated in Fig.~\ref{figure:architecture} (a).
To this end, refer to the Triplet Loss~\cite{7298682}, we set $(\phi(\boldsymbol{v}_b, \boldsymbol{t})-\phi(\boldsymbol{v}_f, \boldsymbol{t}))>m$ as our optimization target for video-level metric learning.
Rather than simply leveraging the margin or triplet-loss based contrastive learning objectiveness, we refer to the recently proposed general pair weighting framework that origins from deep metric learning~\cite{wang2020vitaa,wang2019multi}, which endows our learning objectiveness with the ability of gradient weighting using the logistic-loss as the basic form function. Comparing to the previous works as in~\cite{Mithun_2019_CVPR, chen2020look}, WSRA adaptively assigns proper weights to valuable learning pairs thus benefiting the training. 
We denote fore/back-ground similarity scores as $s_p^i=\phi(\boldsymbol{v}_f^i, \boldsymbol{t}^i)$ and $s_n^i=\phi(\boldsymbol{v}_b^i, \boldsymbol{t}^i)$, respectively. Our video-level loss function is calculated by:
\begin{equation}
    \mathcal{L}_{video} = \log\Big[1+\sum_{i=1}^{N}\exp(\tau(s_n^i-s_p^i+m))\Big],
    \label{eq:video}
\end{equation}
in which $i$ indices the sample in a random mini-batch, $m$ is a predefined margin and $\tau$ is a temperature factor.

\subsection{Snippet-level Attention Modeling}
While the above video-level loss imposed on a whole video encourages learning a discriminative embedding space and the metric functions, we introduce snippet-level modeling to enhance the contrastive study of individual video snippet and multiple textual queries, as illustrated in Fig.~\ref{figure:architecture} (b). Under this case, for the $t$-th snippet in the $j$-th video $\boldsymbol{v}_t^j$ , we calculate the referring weight $\beta_t^i$ as:
\begin{equation}
    \beta_t^j = \frac{\text{exp}(\phi(\boldsymbol{v}_t^j, \boldsymbol{t}^j))}{{\sum_{i=1}^{N}\text{exp}(\phi(\boldsymbol{v}_t^j, \boldsymbol{t}^i))}}.
    \label{eq:snippetweight}
\end{equation}
Since different textual queries don't have such continuity among different snippets, instead of synthesizing textual features, we use this referring weight as penalty on the optimization target. Denoting $\boldsymbol{t}^i$ as the textual query for the $i$-th video and $\boldsymbol{t}^j$ for the  $j$-th video, we define out penalized optimization target as $(\beta_t^j\phi(\boldsymbol{v}_t^j,\boldsymbol{t}^i)-\beta_t^i\phi(\boldsymbol{v}_t^j,\boldsymbol{t}^j))>m$. This target is more flexible during training because those less-optimized scores will have larger weighting factors and consequentially get larger gradient. Our snippet-level loss function for each video is derived as follows:
\begin{equation}
    \mathcal{L}_{snippet} = \log\Big[1+\sum^{T}_{t=1}\exp(\tau(\beta_t^+ s_n^t - \beta_t^- s_p^t+m))\Big],
\end{equation}
where $s_p^t=\phi(\boldsymbol{v}^t, \boldsymbol{t}^+)$ and $s_n^i=\phi(\boldsymbol{v}^t, \boldsymbol{t}^-)$ are the similarity scores of snippet feature $\boldsymbol{v}^t$ with $\boldsymbol{t}^+$ (the ground-truth referring text) and $\boldsymbol{t}^-$ (the sampled negative text), respectively.

\begin{figure*}[t]
\begin{center}
    \includegraphics[width=.92\textwidth]{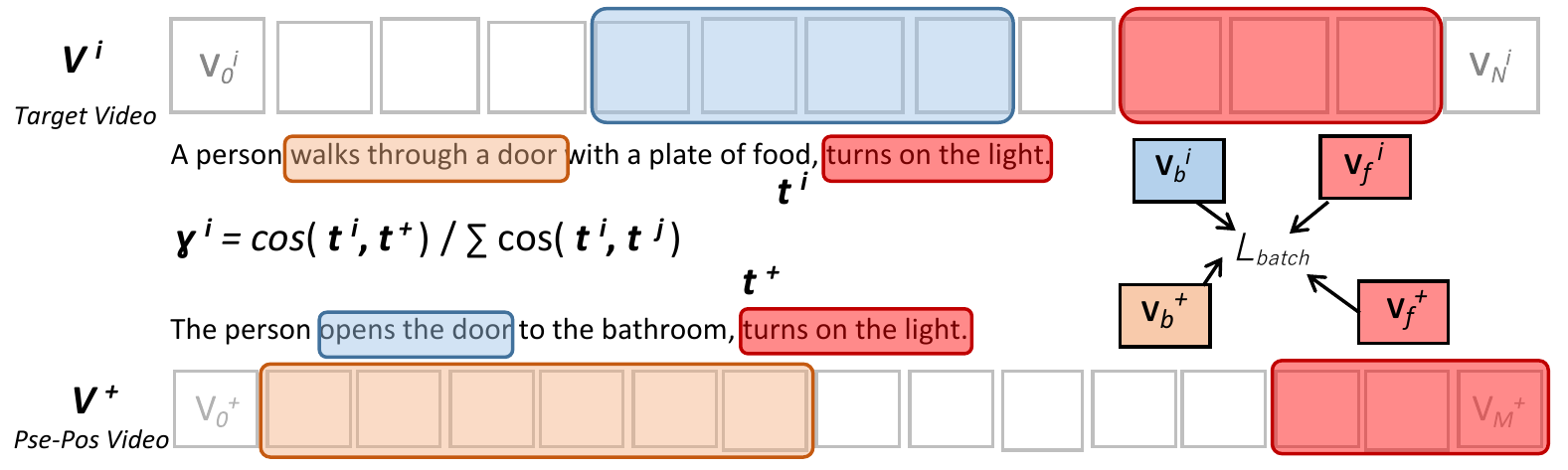}
\end{center}
\vspace{-6mm}
\caption{\small Our batch-level attention modeling is constructed upon cross-video similarity learning: it leverages the fore/back-ground representations as the learning target across samples when similar visual content occurs. We utilize the cosine similarities of the textual embedding as the quantizer for generating pseudo-positive pairs. Note that, the textual descriptions for computation can be either complete sentences or parsed verb/noun phrases, the choice should be hinged on the properties of specific datasets.
}
\label{figure:crossvideo}
\end{figure*}

\begin{figure}[h]
\begin{center}
    \includegraphics[width=.49\textwidth]{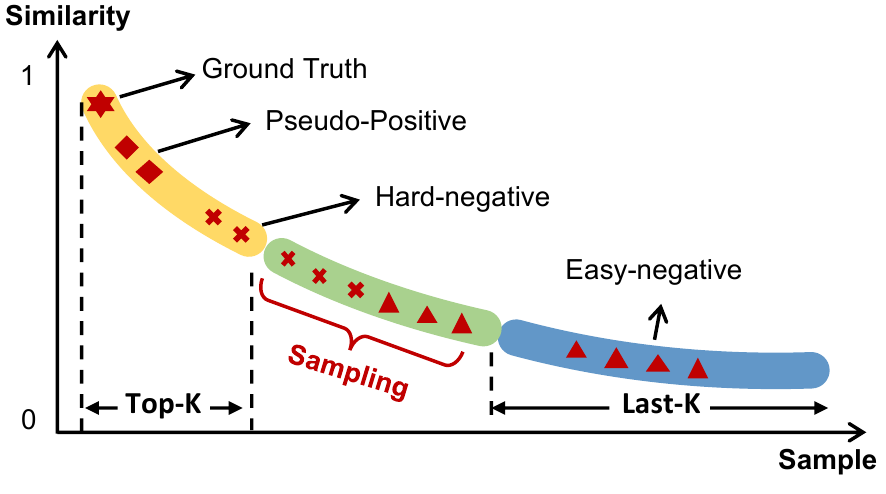}
\end{center}
\vspace{-6mm}
\caption{\small Demonstrative figure of queries' semantic similarity within a single training batch. The $y$-axis stands for distance measured by matching scores between
all queries and the current query in question;
$x$-axis sorts all the queries according the  matching score 
within this training batch. This sampling notion is the basis for our batch-level attention modeling for constructing valid cross-videos positive samples.
}
\label{figure:topksample}
\end{figure}

As noted, sampling semantically similar queries affect learning~\cite{wu2018unsupervised}, but when the batch size grows with limited number of textual expressions, a single training batch may contain semantically similar queries more easily, \emph{e.g.}, ``\textit{a man having food}'' and ``\textit{the man eating dinner}''. 
As a result, simply treating all other textual queries within the mini-batch as the negative samples hurt discriminative learning.
We provide our solution as below on using the similarity $(\boldsymbol{t}^i, \boldsymbol{t}^j)$ to differentiate individual instances (queries within a single
training batch), telling whether a pair of them are semantically dissimilar enough
to support a sampled negative query.

\subsubsection{Top/last-$K$ Sampling.} \label{sec:sampling} 
Inspired by the hard negative sampling techniques in metric learning~\cite{shrivastava2016training,lin2017focal,liu2016ssd,zhang2017range}, 
we propose a top/last-$K$ sampling strategy akin to semi-hard negative sampling~\cite{schroff2015facenet}. 
We identify queries as ``pseudo-positive samples'' and ``easy-negative samples'', 
as demonstrated by Fig.~\ref{figure:crossvideo},
then we select the ``hard-negative samples'',
which provide more informative gradients that help learning.
We note ``easy-negative samples'' do not contribute to better training while being included do not hurt either yet
waste wall-clock time in computation.
To identify the ``hard-negative samples'', 
we first sort all the queries in the mini-batch based on 
the similarity scores
compared to the one given for the video of interest.
Then we simply remove the top and last $K$ samples as easy positive and negative ones.
While the hyper-parameter $K$ depends on the batch size,
we set $K=3$ in our experiments and find it quite 
stable and beneficial to the training.

\subsection{Batch-level Attention Modeling}
Proceeding from previous efforts in un/weak-supervised video representation learning, we are inspired with the necessity to construct contrastive learning pairs across video samples.~\cite{paul2018w} proposed to encourage the videos containing identical actions to be encoded as similar features in their corresponded temporal regions.~\cite{dwibedi2019temporal} learns the frame-wise correspondence across videos and yields great representations without any strong supervision. These works all unanimously highlight that mining the common-information among samples could enormously benefit the discriminative learning. This practice also applies for our video and language tasks, where the similar visual content can be observed in different videos. For instance, the excluded pseudo-positive samples from our snippet-level learning might be ``\textit{a gentleman is having dinner}'', whose visual foreground should be similar with the scenario of ``\textit{man eating food in a restaurant}''. 

Concretely, we assume that for each target video $\boldsymbol{v}^i$, it contains implicit common activities with the video $\boldsymbol{v}^+$ that related to its textual pseudo-positive sample. 
Therefore, we exploit a batch-level attention mechanism as well as an inter-video loss to further improve the discriminative learning by utilizing the mined pseudo-positive samples as stated previously. Following the setting in Sec.~\ref{sec:sampling}, we define the attention weight as:
\begin{equation}
    \gamma^i = \frac{\text{exp}(\cos(\boldsymbol{t}^i, \boldsymbol{t}^+))}{{\sum_{j=1}^{N}\text{exp}(\cos(\boldsymbol{t}^i, \boldsymbol{t}^j))}}.
    \label{eq:batchweight}
\end{equation}
where $\boldsymbol{t}^+$ is the pseudo-positive sample for the textual query $\boldsymbol{t}^i$ that corresponding to the $i$-th video. Here we employ the synthesized fore/back-ground features in Eq.~\ref{eq:foreback}. We use the form of penalized optimization target as $\gamma^i(\cos(\boldsymbol{v}_b^i,\boldsymbol{v}_f^+)-\cos(\boldsymbol{v}_f^i,\boldsymbol{v}_f^+))>m$. Our learning objective to be making their foreground representations closer and enlarging the divergence of foreground with background as is shown in Fig.~\ref{figure:crossvideo}. The overall batch-level loss is expressed as:
\begin{equation}
    \mathcal{L}_{batch} = \log\Big[1+\sum_{i=1}^{N}\exp(\tau\gamma^i(s_n^i-s_p^i+m))\Big].
    \label{eq:batch}
\end{equation}
Since the comparing targets are from the same modality, we simply use cosine similarity in the calculation, that $s_p^i=\cos(\boldsymbol{v}_f^i, \boldsymbol{v}_f^+)$, and $s_n^i=\cos(\boldsymbol{v}_b^i, \boldsymbol{v}_f^+)$.

\subsection{Overall Training Loss}
As for the overall objective loss,
we combine all the above loss terms for end-to-end training our model:
\begin{equation}
    \mathcal{L} = \alpha\mathcal{L}_{video} + \beta\sum^{T}\mathcal{L}_{snippet} 
    + \delta\mathcal{L}_{batch} \\
    \label{eq:overall_loss}
\end{equation}
where $\alpha, \beta$, and $\delta$ are hyper-parameters weighting the loss terms, which are set to 0.1, 1, 0.1, respectively.
We list details on the effect of these terms in the ablation study and further study the effect of various loss weights in the supplementary materials.
Due to the limited computational resources, we empirically set temperature weight as constant 1 and did not conduct exhaustive searching for the optimal temperature weight per loss. However, various values slightly affect the performances.
During training, we use the Adam optimizer with constant learning rate 1$\times10^{-4}$ and coefficients 0.9 and 0.999 for computing running averages of gradient and its square.
We implement our algorithm using PyTorch toolbox~\cite{paszke2017automatic} on single GTX1080 Ti GPU.

\section{Experiment}
The goal of our experiments is to validate the effectiveness of 
the proposed weakly-supervised framework with the top-down referring attention (\textit{WSRA}) 
for temporal-textual association learning,
through two tasks of moment localization and language grounding on
the  DiDeMo~\cite{hendricks2018localizing} and Charades-STA~\cite{gao2017tall},
respectively.
We use the mean Average Precision (mAP) 
under various Intersection over Union (IoU) thresholds to measure the performance. 
First, we elaborate on some important details about the models and language features.
We then compare our \textit{WSRA} with other state-of-the-art methods with
systematical ablation studies on the two tasks over the two benchmarks.
Finally, we visualize the qualitative results produced by our \textit{WSRA}.

\noindent \textbf{Language Processing and Feature Extraction}.
The proposed  \textit{WSRA} framework is agnostic to the choice of language models.
Although one is free to use any language models providing features to represent
the textual queries (\emph{e.g.}, natural-language sentence or short phrase),   
we turn to the Openai-GPT2~\cite{radford2019language}, 
which is a released language model\footnote{\url{https://github.com/openai/gpt-2}}
trained over large-scale, diverse corpus (Wikipedia, news and books). 
GPT2 is composed of a stacking of repetitive transformer modules, and we retrieve our textual features by averaging all outputs from modules as done in literature~\cite{xiao2018bertservice}.


\subsection{Moment Localization on DiDeMo}

For localizing moments over a given natural-language description,
we use the Distinct Describable Moments (DiDeMo) dataset~\cite{hendricks2018localizing}, 
which includes $>$10k 25-30 second long Flickr videos. 
Manual annotations contain  $>$40k sentences with temporal boundaries 
for fine localization. 
In total, the DiDeMo dataset contains 
8,395, 1,065 and 1,004 videos for training, validation and testing respectively. We report the performances of our model on the testing split and conduct ablation studies on the validation split.
As suggested by the dataset, 
to simplify association between the sentence and a video, 
each long video is divided into 6 segments, which are annotated as whether corresponding to the sentence for localization. 
In DeDeMo, each textual description is associated with only one continuous video moment, thus yielding 21 possible candidates ($\sum_{i=1}^{6}i$). 
We measure the performance using the Rank@1 (R@1) and Rank@5 (R@5) (accuracy of the top-1/5 retrieved candidates) and their mean Intersection of Unions (mIoU) when IoU=1. To extract visual features of the videos, 
we use the official provided features 
over both RGB and optical flow~\cite{anne2017localizing}. 
For fair comparisons with other methods, as done in~\cite{hendricks2018localizing,anne2017localizing}, we compute the visual features as the concatenation of the global averagely pooled features (all 21 proposal features) and the local proposal features, which are produced by after average pooling the features from each segment.
We set hyper-parameters $\beta=1$ and $\alpha=\delta=0.1$ in this experiment and report results under different combinations in our supplementary materials.
During inference, we select the top-5 proposals with highest attention weights as the final prediction.


{
\setlength{\tabcolsep}{0.14em} 
\begin{table}[t]
\captionsetup{font=small}
\centering
\small
\begin{tabular}{ccc|ccc|ccc}
\multirow{2}{*}{$\mathcal{L}_{video}$} & \multirow{2}{*}{$\mathcal{L}_{snippet}$}  & \multirow{2}{*}{$\mathcal{L}_{batch}$} & \multicolumn{3}{c|}{DiDeMo (Val)} & \multicolumn{3}{c}{Charades-STA} \\
\cline{4-9}
 &  & & R@1 & R@5 & mIoU & R@1 & R@5 &mIoU\\
\hline


\checkmark & -- & --  & 10.10 & 36.05  & 18.78  & 3.42 & 18.56 & 16.85\\
-- & \checkmark & -- & 14.68  & 45.72  & 26.04 & 4.84 & 23.65 &22.16 \\
\checkmark & \checkmark & --  & {16.20} & 49.56 & {27.50} & 8.82 & 26.65 & 26.42 \\
\checkmark & \checkmark  & \checkmark  & \textbf{16.92}  & \textbf{50.12}  &  \textbf{28.32} & \textbf{11.01} &  \textbf{39.02} & \textbf{31.00}

\end{tabular}
\caption{Ablation study of loss terms on the validation split of DiDeMo and Charades-STA when IoU is 1 and 0.7 respectively. As the $\mathcal{L}_{batch}$ does not explicitly incorporate the alignment learning, we study it as the complementing term \textit{w.r.t.} the first two terms.
}
\label{table:ablation_studies}
\end{table}
}

{
\setlength{\tabcolsep}{0.1em} 
\begin{table}[t]
\small
\begin{center}
\centering
\begin{tabular}{c|c|c|ccc}
\multirow{1}{*}{Supervision} & \multirow{1}{*}{Method} & \multirow{1}{*}{Feature} &
R@1 & R@5 & mIoU  \\
\cline{1-6}

& Upper Bound & -- & 74.75 &  100 & 69.05  \\
& Chance & --  & 3.75 & 22.5 & 22.64  \\
\cline{2-6} 
\multirow{6}{*}{\shortstack{\textit{Fully}\\Supervised}} & CCA~\cite{anne2017localizing} & Flow\&RGB & 18.11  & 52.11 & 37.82  \\
& Lang. Obj. Retr.~\cite{Hu_2016_CVPR} & {Flow}  & 16.20 & 43.94 & 27.18  \\
& LSTM-RGB-local~\cite{anne2017localizing} & {RGB}  & 13.10 & 44.82 & 25.13  \\
& LSTM-Flow-local~\cite{anne2017localizing} & {Flow}  & 18.35 & 56.25 & 31.46  \\
& MCN~\cite{anne2017localizing} & {Flow\&RGB}  & 28.10 & 78.21 & 41.08 \\
& TGN~\cite{chen2018temporally} & {Flow\&RGB}  & \textbf{28.23} & \textbf{79.26} & \textbf{42.97} \\
\hline
\multirow{5}{*}{\shortstack{\textit{Weakly}\\Supervised}}
& TGA~\cite{Mithun_2019_CVPR} & {Flow\&RGB}  & 12.19 & 39.74 & 24.92 \\
& WSRA  & {RGB}  & 14.20 & 43.67 & 25.22 \\
& WSRA$^{*}$ & {Flow}  & {17.23} & 48.84 & {27.42} \\
& WSRA & {Flow}  & \textbf{17.88} & 50.04 & \textbf{29.90} \\
& WSRA & {Flow\&RGB}  & 17.52 & \textbf{52.11} & 28.87 \\
\end{tabular}
\end{center}
\caption{\small Comparison of performances with fully/weakly-supervised methods on DiDeMo test split. Our \textit{WSRA} model outperforms the state-of-the-art weakly-supervised method (TGA)~\cite{Mithun_2019_CVPR}. We report performances of models using regular word vectors (marked as \textit{WSRA}$^{*}$). ``Chance'' denotes the results of random guess.
}
\label{table:didemo}
\end{table}
}

{\setlength{\tabcolsep}{.5em} 
\begin{table*}[t]
\begin{center}
\small{
\begin{tabular}{c| c | cc | cc | cc|c}
\multirow{2}{*}{} &  \multirow{2}{*}{Approach} & \multicolumn{2}{c}{\textbf{IoU=0.3}}  & \multicolumn{2}{c}{\textbf{IoU=0.5}}  & \multicolumn{2}{c}{\textbf{IoU=0.7}}\\
 & & R@1& R@5 & R@1  & R@5   & R@1 & R@5 &  mIoU \\
\hline

\multirow{5}{*}{{\shortstack{\textit{Fully}\\Supervised}}}&Random~\cite{gao2017tall} & -- & -- & 08.51 & 37.12  & 03.03 & 14.06 & --\\
&VSA-STV \cite{kiros2015skip} & -- & --  & 10.50 & 48.43  & 04.32 & 20.21 & --\\

&CTRL~\cite{gao2017tall} & -- & -- & 21.42 & 59.11 & 07.15 & 26.91 & -- \\
&Xu \emph{et al.} ~\cite{xu2019multilevel} & -- &-- & 35.60 & 79.40  & 15.80 & 45.40 & -- \\
&MAN \cite{zhang2019man} & -- & -- & 46.53 & 86.23  & 22.72 & 53.72 & --\\

\hline 
\multirow{4}{*}{{\shortstack{\textit{Weakly}\\Supervised}}}&TGA~\cite{Mithun_2019_CVPR} & 32.14 & {89.56} & 19.94 & {65.52} & 08.84 & 33.51 & --\\
&SCN~\cite{lin2019weakly} &  42.96 & \textbf{95.56}  & 23.58 & \textbf{ 71.80}& 09.97 & {38.87} & --\\
&CTF$^{*}$~\cite{chen2020look} &  39.80 & - &  27.30 & - & \textbf{12.90} & - & --\\
&WSRA (Ours) & \textbf{50.13} & {86.75}  & \textbf{31.20} & 70.50& {11.01} & \textbf{39.02 }& \textbf{31.00} \\
\end{tabular}
}
\end{center}
\vspace{-3mm}
\caption{ Language grounding results on the test set of Charades-STA under different intersection over unions. $^*$ denotes the work under peer reviewing.
\label{table:charades_sta}
}
\end{table*}
}

\begin{figure}[t]
\includegraphics[width=.45\textwidth]{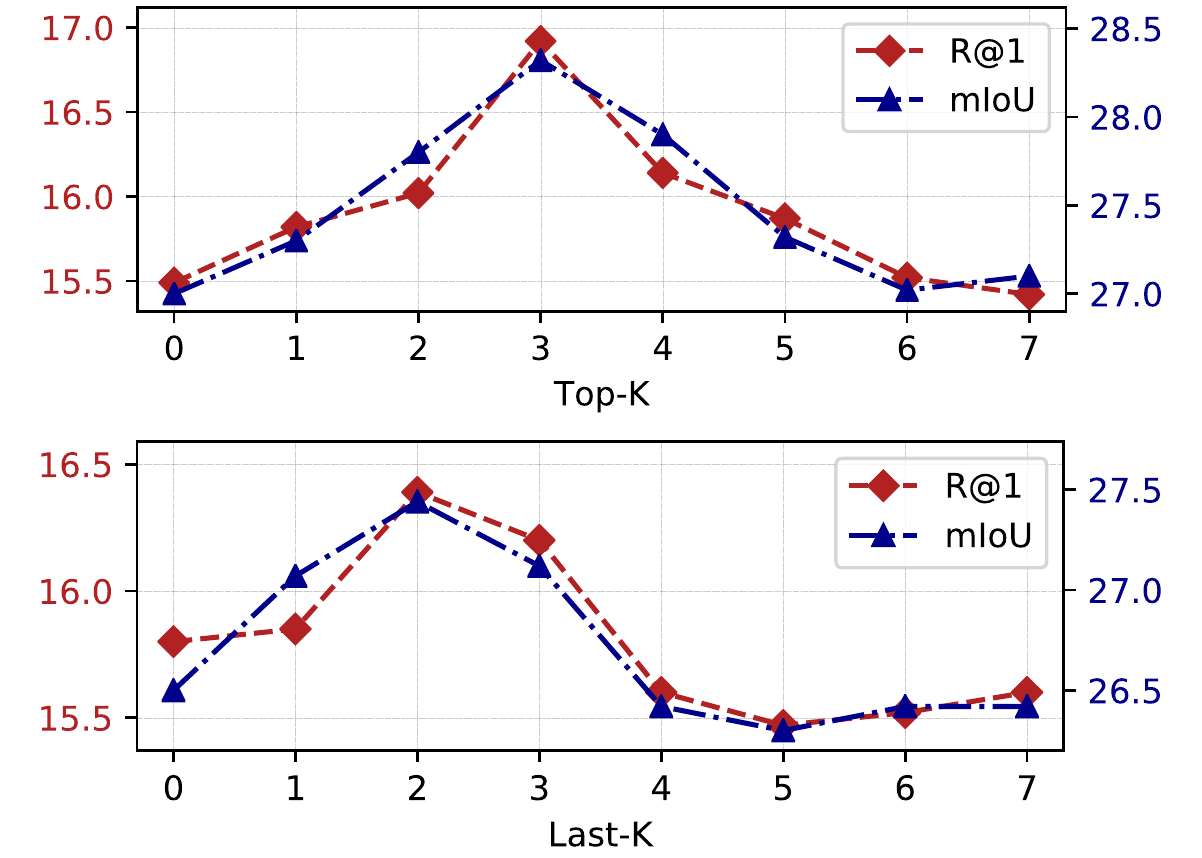}
\centering
\vspace{-3mm}
\caption{\small  Effect of different $K$ in the top/last-$K$ sampling on DiDeMo validation split.}
\label{figure:k_sample}
\end{figure}

To understand how each loss term contributes to the performance,
we conduct a systematical ablation study in Table~\ref{table:ablation_studies}. 
The results clearly demonstrate that all the loss terms improve the performance individually and are complementary with each other, and by combining them all achieves the best performance. Since $L_{batch}$ supervises only the visual features without any alignment, thus can not be compared independently.
From this table, it is worth noting that using the fore/background loss only performs on par with TGA~\cite{Mithun_2019_CVPR} (see Table~\ref{table:didemo}), which is the weakly-supervised method that adopts similar design of fore/back-ground modelling. 
Our final \textit{WSRA} outperforms TGA~\cite{Mithun_2019_CVPR} by a clear gain,
demonstrating that the loss terms  exploits the  weak supervision more
effectively in learning a better discriminative model.
Fig.~\ref{figure:k_sample} studies the effect of our 
top/last-$K$ sampling strategy with various $K$. 
We clearly see that, when excluding top $K=3$ and last $K=2$ samples in the mini-batch, it yields the best performance,
with the batch size set to 42 in this experiment.

We compare our \textit{WSRA}  with other advanced weakly/fully-supervised methods in Table.~\ref{table:didemo}. As the previous methods, 
we experiment with visual features from modals of RGB and optical flow. Upper-bound of DiDeMo is brought since the human annotators cannot achieve 100\% agreement in annotating the segment boundaries \emph{w.r.t} the given video-level sentence~\cite{anne2017localizing}.
Comparing with the baseline models, our \textit{WSRA} model  significantly outperforms weakly supervised TGA~\cite{Mithun_2019_CVPR} by 5.5\%, 11\% at R@1 and R@5 respectively, even achieving comparable performance with several fully supervised methods \emph{e.g.}, CCA~\cite{anne2017localizing} and LSTM based model~\cite{anne2017localizing}. 
It is also worth noting that, our \textit{WSRA} model has similar number of parameters with all above mentioned baseline models. 

\subsection{Language Grounding on Charades-STA}
We further validate our \textit{WSRA} for the language grounding task on another video dataset, 
Charades-STA~\cite{gao2017tall},
which augments the Charades dataset~\cite{actorobserver,sigurdsson2017asynchronous}  with
manual annotations in the form of 
natural language descriptions at precise start-end timestamp of each video.
Charades-STA contains 12,408/3,720 video-query pairs for training/testing. 
For annotation,
Charades-STA expands each verb to generate a textual sentence
from single caption in Charades using language templates, 
and associate the sentence with frames corresponding to this verb.
To encode the video segment,
as done by other weakly-supervised methods 
for action localization~\cite{nguyen2018weakly,paul2018w},
we use the I3D feature from a pre-trained model trained over 
the Kinetics dataset~\cite{carreira2017quo}. 
For inference, we turn to the moment selection methods~\cite{gao2017tall},
which generate multi-scale sampled moment candidates using sliding window method for retrieval with fixed length of frames. 
Moreover, 
since the  video duration varies significantly, we sampling the moment 
candidates with lengths proportioned to the video duration.
Specifically, sampled candidate clips are with $\small\{$20\%, 30\%, 40\%, 50\%$\small\}$ of the whole videos and in 80\% overlap using the sliding window manner, then the moment with highest attention weight score is selected as the final prediction. More details can be found in the supplementary.
\begin{figure*}[t]
\includegraphics[width=.98\textwidth]{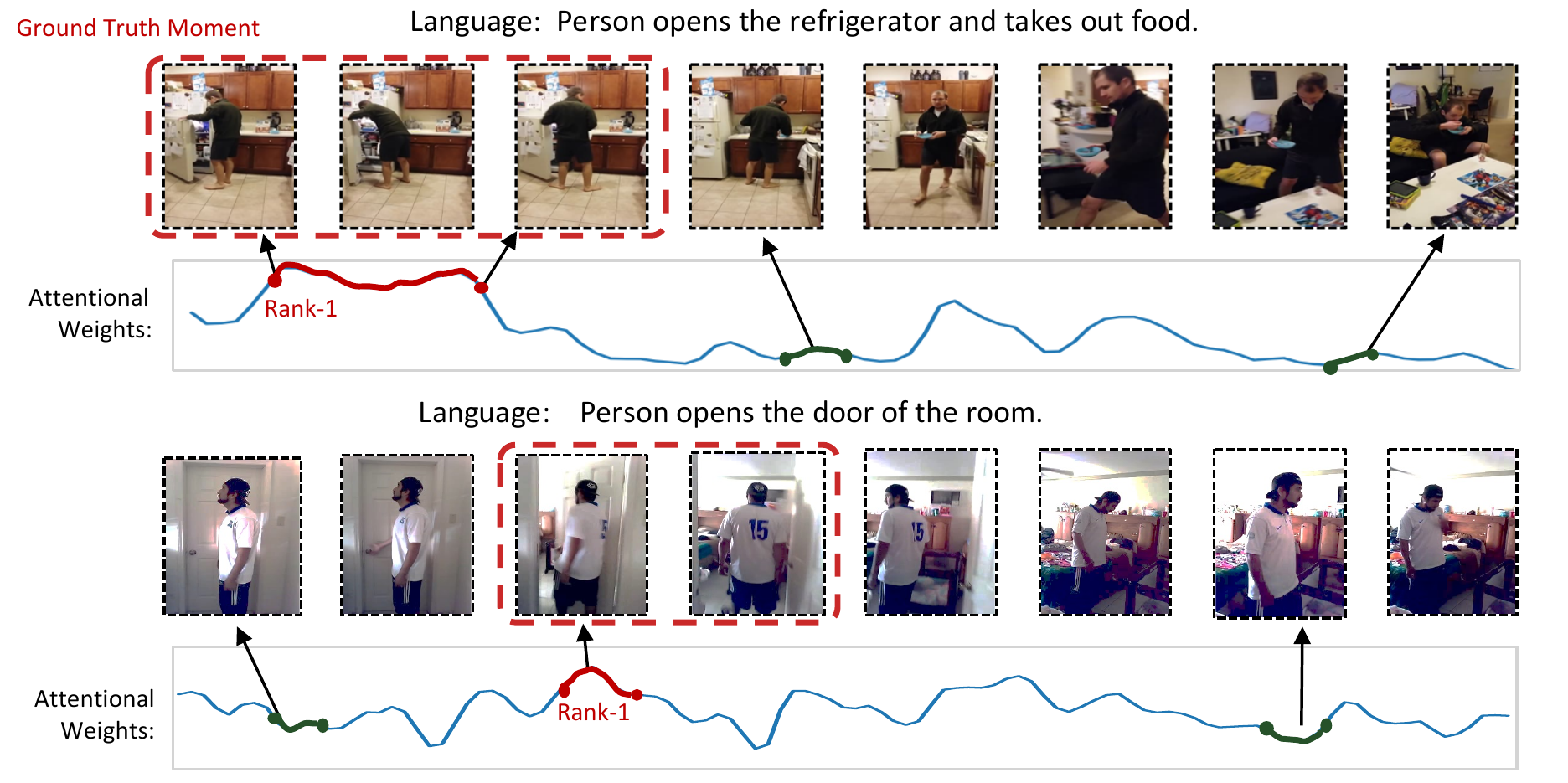}
\centering
\caption{\small Qualitative examples of the language grounding on Charades-STA dataset. We select the top several video moments with high attention weights as the prediction.}
\label{figure:qualitative}
\end{figure*}
We report the performance comparison with
different mean Intersection-over Union (mIoU = \{0.3, 0.5, 0.7\}) and Recall@\{1, 5\}, 
and the mIoU as a summary metric.
We list detailed comparison in Table~\ref{table:charades_sta}.
As TGA~\cite{Mithun_2019_CVPR} is also constructed among the contrastive learning, we study their performances gap and can see the clear advantage of our referring attention:  
\textit{WSRA} significantly outperforms TGA, 
for example by 18\% at [R@1, IoU=0.3], 
and 12\% at [R@1, IoU=0.5], respectively. 
We can also see clearly from the above table that, comparing with other state-of-the-art weak-supervised methods, \textit{WSRA} demonstrates clear leading at R@1 when IoU=0.3 and 0.5. Also, \textit{WSRA} demonstrates either comparable or even better performance than 
most fully-supervised methods.
In the meanwhile, our top/last-$K$ sampling also rejects theses pseudo-positive and less informative samples in the contrastive learning. Fig.~\ref{figure:qualitative} shows the qualitative examples of grounding in Charades-STA dataset where the moment is aligned with language query even facing much longer videos. We further study the effect of cross-video loss in the supplementary.
\section{Conclusion and Broader Impact}
In this paper, we propose the weakly supervised model with the referring attention mechanism (\textit{WSRA})  for learning temporal-textual association on videos.
We introduce several novel loss terms and sampling strategies,
all of which help better learning by fully exploiting the cues from 
the weak supervision at video level.
Through extensive experiments on two benchmarks,
we show the \textit{WSRA} model outperforms the state-of-the-art weakly-supervised
methods by a notable gain,
achieving on par or even better performance than some fully-supervised methods. As an outlook for the future study, we consider that the most potential aspect our model would benefit comes from the video-language representation learning at scale~\cite{miech2019howto100m}, whereas the training is often severely accompanied by uncurated annotations: \textit{e.g.}, temporally misaligned descriptions~\cite{miech2019end}. To construct a soundly and largely pre-trained model, it is requisite to properly leverage the weak or biased annotation at comprehensive views. Our \textit{WSRA} provides us with such a perspective as the trailblazer, that investigates thoroughly how language can be fully exploited as valid supervisions even without temporal annotations.




\end{document}